\documentclass[journal]{IEEEtran}

\ifCLASSINFOpdf
  \usepackage[pdftex]{graphicx}
  \graphicspath{{./Figures/}}
  \usepackage{epstopdf}
\else
\fi
\usepackage{tikz}
\usetikzlibrary{shapes,arrows,decorations.pathreplacing, calc}

\usepackage{amsmath}
\usepackage{amssymb}

\usepackage{bbold}

\usepackage{array}
\usepackage{multirow}

\ifCLASSOPTIONcompsoc
 \usepackage[caption=false,font=normalsize,labelfont=sf,textfont=sf]{subfig}
\else
 \usepackage[caption=false,font=footnotesize]{subfig}
\fi

\usepackage{paralist}
\usepackage{adjustbox}
\usepackage{makecell}
\usepackage{setspace}
\usepackage{blindtext}

\newcommand{\etal}{\textit{et al}.}

\usepackage{algorithm}      % algorithm\usepackage{graphicx}
\usepackage[noend]{algpseudocode} % algorithm

\newcommand{\algrule}[1][.2pt]{\par\vskip.1\baselineskip\hrule height #1\par\vskip.1\baselineskip}
\algdef{SE}[DOWHILE]{Do}{doWhile}{\algorithmicdo}[1]{\algorithmicwhile\ #1}%

\begin{document}

% \title{Doppler Centroids Estimation on Subapertures for Physic Guided Ocean SAR Image Retrieval}
\title{Guided Unsupervised Learning by Subaperture Decomposition for Ocean SAR Image Retrieval}

\author{Nicolae-C\u{a}t\u{a}lin Ristea,
        Andrei Anghel,~\IEEEmembership{Senior Member,~IEEE,}
        Mihai Datcu,~\IEEEmembership{Fellow,~IEEE},
        Bertrand Chapron
        
\thanks{

\noindent N.C. Ristea and A. Anghel are with the Research Centre for Spatial Information (CEOSpaceTech) and the Department of Telecommunications, University Politehnica of Bucharest, Romania.\\
M. Datcu is with the Research Centre for Spatial Information (CEOSpaceTech), University Politehnica of Bucharest, Romania and the Remote Sensing Technology Institute, German Aerospace Center (DLR), Germany.\\
B. Chapron is with Laboratoire d’Ocanographie Physique et Spatiale (LOPS), Ifremer, Brest, France.}
\thanks{}
}

\markboth{Journal TGRS, September~2022}%
{Shell \MakeLowercase{\textit{et al.}}: Bare Demo of IEEEtran.cls for Journals}

\maketitle

\begin{abstract}
Spaceborne synthetic aperture radar (SAR) can provide accurate images of the ocean surface roughness day-or-night in nearly all weather conditions, being an unique asset for many geophysical applications. Considering the huge amount of data daily acquired by satellites, automated techniques for physical features extraction are needed. Even if supervised deep learning methods attain state-of-the-art results, they require great amount of labeled data, which are difficult and excessively expensive to acquire for ocean SAR imagery. To this end, we use the subaperture decomposition (SD) algorithm to enhance the unsupervised learning retrieval on the ocean surface, empowering ocean researchers to search into large ocean databases. We empirically prove that SD improve the retrieval precision with over 20\% for an unsupervised transformer auto-encoder network. Moreover, we show that SD brings important performance boost when Doppler centroid images are used as input data, leading the way to new unsupervised physics guided retrieval algorithms.
\end{abstract}


\IEEEpeerreviewmaketitle

\section{Introduction}
\IEEEPARstart{E}{arth} observation (EO) is the integration of information about planet Earth's physical, chemical and biological system by remote sensing (RS) technologies provided by earth surveillance techniques, including the collection analysis and presentation of data \cite{Jiao-JSTAR-2015}. Considering that the ocean accounts for about 71\% of Earth’s surface, the ocean observation increasingly draws the attention of research community over the last decades. Humans had minimal ocean observations before 1978, when Seasat, the first Earth-orbiting satellite designed for remote sensing of Earth’s oceans was launched \cite{Stewart-AMS-1988}. Although Seasat only operated for about 100 days, the mission acquired more data about the ocean than all previous sensors combined. This event stimulated the fast development of ocean-satellite, leading to a growing number of satellites carrying different sensors (e.g., microwave, visible, infrared) being launched to improve our understanding about the ocean.

Nowadays, one of the most used space-borne sensors for ocean observation is the synthetic aperture radar (SAR), used by satellite mission Sentinel-1 from 2014, when the WV mode was implemented. The WV modality is available only on the Sentinel-1A/B and is dedicated for retrieving ocean surface properties at global scale \cite{Stopa-2016-GRL}. The WV measurements have a spatial resolution of approximately $4$ meters and a scene footprint of $20$ by $20$ km. These sensors collect monthly nearly $120,000$ WV vignettes of the global ocean surface. Moreover, tens of satellites have also been approved for the next 20 years, conducting to a sharp rise of ocean data. Hence, automated systems designed to interpret, extract and find features in big data environments are highly needed to exploit all the available information.

An important aspect of ocean big data is that, having more data does not guarantee more valuable information extracted. Usually, the key information is sparsely hidden in massive ocean-satellite data. Once with the growing capacity of collecting ocean data, many efforts have been put into developing and validating retrieval algorithms to generate standard time series global ocean parameters \cite{Jiao-JSTAR-2015, Espinoza-TGRS-2013, Sumbul-TGRS-2022, Yang-TGRS-2011, Zheng-TGRS-2018, Zheng-TGRS-2018, Li-NSR-2020}. Currently, one important focus is to develop efficient and intelligent approaches to improve the information extraction capability with powerful deep learning algorithms. Because the physical phenomena which can occur on the ocean are diverse, ranging from waves and algal blooms, which are locally generated and their signatures only consist of a tiny percentage of an ocean vignette, to long time series data (e.g., level of ocean), new data-driven information mining algorithms are required. Moreover, extracting real-time information from high-rate downlink satellite data stream requires high-speed data processing. Deep learning techniques can satisfy all mentioned requirements, proving high efficiency and generalization capacity in image related tasks \cite{Dubey-TCSVT-2021, Neyshabur-NIPS-2017}.

Other major aspect of ocean big data is the costly process of annotating data. Considering the particularities of remote sensing ocean data, only people with expertise can annotate vignettes (e.g., ocean currents direction, waves height, ocean phenomena), making the process time consuming and costly. Inspired from visual data domain, several works leverage unsupervised information to learn deep representations \cite{Ye-2019-GRSL, Tang-2018-RS, Jiao-2015-JSTAR, Liu-2020-TGRS}. Nevertheless, even if there is a moderate success on SAR unsupervised image retrieval, there are no works which studies the benefits of unsupervised deep learning (UDL) for ocean SAR image retrieval. 

% To avoid costly annotation, several works leverage unsupervised information to learn deep representations. Unsupervised learning can be used to initialize network weights, as
% in Erhan et al. (2009, 2010). Methods that directly use unsupervised weights include domain transfer (Donahue et al.2014) and k-means (Coates and Ng 2012). Most recently,
% some works have looked into using temporal coherence as
% supervision (Goroshin et al. 2015, 2014). Somewhat related
% to our work, Agrawal et al. (2015) propose to train a network
% by learning the affine transformation between synchronized
% 123
% 152 Int J Comput Vis (2017) 121:149–168
% image pairs for which camera parameters are available.
% Similarly, Jayaraman and Grauman (2015) uses a training
% objective that enforces for sequences of images that derive
% from the same ego-motion to behave similarly in the feature
% space. While these two works focus on a weakly supervised
% setting, we focus on a fully unsupervised one.

In our paper, we extend the previous work from \cite{Ristea-2022-IGARSS} and address the unsupervised ocean image retrieval task by combining the subaperture decomposition (SD) algorithm with UDL. Using UDL for image retrieval we exclude the necessity of labeled data, and combining it with the preprocessing algorithm based on SD, we pushed the retrieval accuracy closer to the supervised learning approach. Moreover, we tested our approach for a physics guided remote sensing approach (providing as input to the network Doppler centroid images of the vignettes), and we obtained important improvements when using SD beforehand.
Using those processing techniques, we developed an efficient algorithm of query by image, meant to help experts to identify similar phenomena on the ocean surface. Each vignette is described by an embedding vector computed with a pretrained deep neural network (DNN), trained in an unsupervised manner. Moreover, we extended the use case of query by image to a more complex approach of query by physical parameters. More exactly, we estimated the Doppler centroid images of the subaperture single-look complex (SLC) vignettes and used them as inputs of the DNN. In this case each vignette is described by an embedding vector, taken from a DNN pretrained on Doppler centroid images estimated on subapertures.

In summary, with respect to our previous work \cite{Ristea-2022-IGARSS}, our current contribution is twofold:
\begin{itemize}
    \item  We are the first who proposed an unsupervised query by example framework for ocean SAR imagery.
    \item  We combined the previous SD algorithm with DCE and obtained superior results for classification and image retrieval.
\end{itemize}

\section{Related Work}
This section makes a state-of-the-art analysis relative to the proposed methodology and covers the following topics: subaperture decomposition in SAR imagery, Doppler centroid estimation methods, transformer models and image retrieval techniques.
\begin{figure*}[!t]
\begin{center}
\centerline{\includegraphics[width=0.95\linewidth]{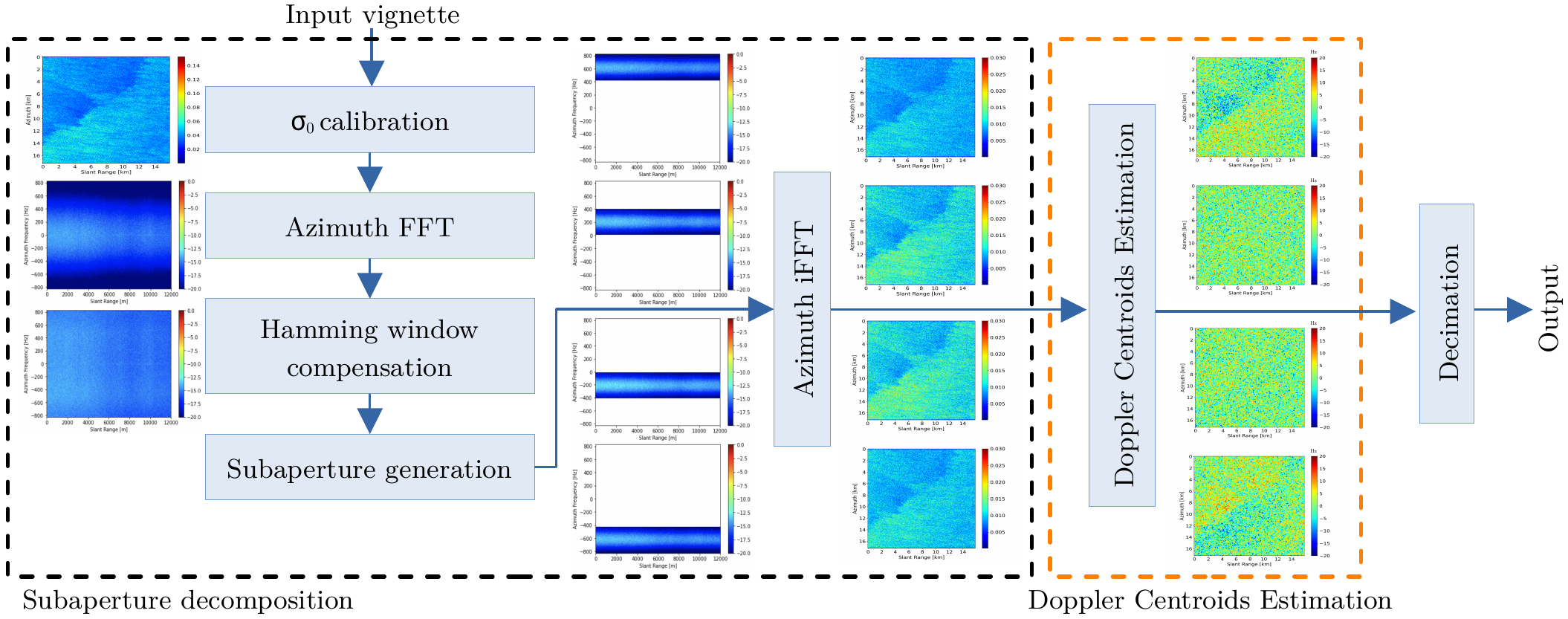}}
% \vskip -0.3cm
\caption{The preprocessing subaperture decomposition pipeline. An input vignette is processed by a series of blocks, followed by the subaperture generation. Next, each subaperture is processed by the azimuth inverse FFT. The Doppler Centroid Estimation block is an additional algorithm, which can be omitted in accordance with the experiment performed. The output is either the decimated subapertures or the decimated Doppper centroids on the subapertures.}
\label{fig_pipeline}
\end{center}
% \vskip -1.0cm
\end{figure*}

\subsection{Subaperture Decomposition}

The SD algorithm is widely used for SAR imagery \cite{Wang-2020-GRSL, Brekke-2013-GRSL, Singh-2010-IGARSS, Focsa-2020-COMM, Lu-RSTA-2011, Ristea-2022-IGARSS}. The method was combined with both classical signal processing algorithms \cite{Brekke-2013-GRSL, Singh-2010-IGARSS, Focsa-2020-COMM} and deep learning techniques \cite{Wang-2020-GRSL, Ristea-2022-IGARSS}. In \cite{Brekke-2013-GRSL} the SD is proposed for the ship detection task, while in \cite{Singh-2010-IGARSS} it is used for target characterization. Moreover, the SD algorithm was used to translate a single channel SAR image into three channels image alike representation, by decomposing it into three sub-bands. Afterwards, the authors used pretrained DNN for target classification task on the ground \cite{Wang-2020-GRSL}. 

Distinctly, we propose to extend the SD usage from our previous work \cite{Ristea-2022-IGARSS} by combining the SD with unsupervised learning to improve the ocean image retrieval. To the best of our knowledge, we are the first who use SD in an unsupervised deep retrieval algorithm. Moreover, we use the SD to improve the classification and unsupervised retrieval for the DCE algorithm.

\subsection{Doppler Centroids Estimation}

For decades Doppler centroids are used in processing SAR data \cite{Raney-IJRS-1986, Madsen-TAES-1989}. Many works have been proposed to improve the Doppler estimation in specific settings \cite{Wong-TGRS-1996, Lopez-TGRS-2008, Zhang-JSTARS-2021}. In \cite{Wong-TGRS-1996} authors expose an end-to-end Doppler centroid estimation scheme, which resolves the Doppler ambiguity and works on various terrain types, including land, water and ice, while in \cite{Lopez-TGRS-2008} the authors discuss temporal and phase synchronization for bistatic SAR and the Doppler estimation procedure. Hansen \etal \cite{Hansen-TGRS-2011} presented the processing steps and error corrections needed to retrieve estimates of sea surface range Doppler velocities from ENVISAT advanced SAR wide swath medium resolution image products. They addressed the retrieval accuracy based on examination of the corrected Doppler shift measurements. 

Mainly, DCE approach was used in various SAR processing chains from focusing algorithms to parameters estimation. Differently, we combine the SD and a DC estimation algorithm (the one proposed in \cite{Madsen-TAES-1989}, but applied on a sliding two-dimensional window) to obtain physics based representations for ocean SAR vignettes. Those representations are used for classification and unsupervised physics guided image retrieval, empirically proving that the SD significantly improves the performance when is used as a preprocessing stage before DCE. This leads the way to new unsupervised physics guided retrieval algorithms.

\subsection{Transformer models}
Due to the recent progress of attention mechanisms \cite{Vaswani-NIPS-2017}, transformers have become attractive and powerful choices for SAR related tasks \cite{Dong-2021-TGRS, Basit-2022-RS, Li-2022-RS}. In \cite{Dong-2021-TGRS} a vision transformer (ViT) \cite{Vaswani-NIPS-2017} based representation learning framework is proposed, which use self-attention to replace
convolution, which shifts the focus from the information in local neighborhoods to the long-range interactions between each pixel. In \cite{Basit-2022-RS} the authors add a gradient profile loss to the classical CNNs and vision transformer based hybrid models for oil spills in SAR imagery. Li \etal propose an enhancement Swin transformer detection network, named ESTDNet, to complete the ship detection in SAR images to solve the issues related to the characteristics of strong scattering, multi-scale, and complex backgrounds of ship objects in SAR images. 

In our work, we aim to benefit from the modeling power of transformers while being able to process reasonable down-sampled ocean SAR images, we adopt a generative convolutional transformer with a manageable number of parameters called CyTran \cite{Ristea-2021-ARXIV}. We used it in an unsupervised set-up, showing that, using SD as a preprocessing stage we improve the SAR image descriptors, leading to an important precision boost for image retrieval.

\subsection{Image Retrieval}

The content based image retrieval aims to find images from a large scale data set, which are similar with a query image. Generally, the similarity between the features of the query image and all others images from the data set is used to rank the images for retrieval. Thus, the performance of any image retrieval algorithm depends on the similarity computation between samples. Ideally, the similarity score between two images should be discriminative, robust and efficient. Various methods based on hand-crafted descriptors \cite{Leng-ACCESS-2018, Koteswara-MSSP-2019, Bedi-PRIA-2020}, distance metric learning \cite{Bellet-ARXIV-2013, Wang-TPAMI-2017, Kaya-MDPI-2019}, deep learning models \cite{Wan-ACMMM-2014, Dubey-TCSVT-2021} and unsupervised learning \cite{Ye-2019-GRSL, Tang-2018-RS, Jiao-2015-JSTAR, Liu-2020-TGRS} have been proposed for the image retrieval task. However, the deep learning has emerged as a dominating alternative of hand-designed feature engineering, the features being learned automatically from data.

More closely to our task, there have been several works for content based image retrieval from remote sensing data \cite{Jiao-JSTAR-2015, Sumbul-TGRS-2022, Espinoza-TGRS-2013, Ye-GRSL-2019, Xiong-JSTARS-2020}. In \cite{Espinoza-TGRS-2013} authors propose a classical approach for EO image retrieval based on enriched metadata, semantic annotations and image content. The solution generates an EO-data model by using automatic feature extraction, processing the EO product metadata and defining semantics, which later is used to answer complex queries. Ye \etal \cite{Ye-GRSL-2019} propose an unsupervised domain adaptation model based on convolutional neural networks (CNNs) to learn the domain-invariant feature between SAR images and optical aerial images for SAR image retrieving. In \cite{Sumbul-TGRS-2022} authors propose a plasticity-stability preserving multi-task learning approach to ensure the plasticity and the stability conditions of whole learning procedure independently from the number and type of tasks. This is achieved by defining two novel loss functions, the plasticity preserving loss and the stability preserving loss. They reported superior results compared with state-of-the-art methods for content based image retrieval. Regarding the unsupervised image retrieval, in \cite{Tang-2018-RS} the authors combine the unsupervised feature learning method based on the bag-of-words with k-nearest neighbours algorithm for text to image SAR image retrieval. Ye \etal \cite{Ye-2019-GRSL} propose an unsupervised domain adaptation model based on CNN to learn the domain-invariant feature between SAR images and optical aerial images for SAR image retrieving. 

Distinct from all mentioned methods, we exploit the SD remote sensing algorithms to improve the performance of unsupervised image retrieval algorithm, by enriching the ocean SAR image descriptive embeddings. Moreover, we are the first which perform physics guided unsupervised image retrieval based on DCE, opening the frontiers for a new research area.

\section{Method}

\subsection{Subaperture decomposition}

A classical SAR system acquires the backscatter returned from irradiated targets in different positions and different azimuth angles along the radar trajectory. The real antenna aperture is replaced by the synthetic aperture to obtain high azimuth resolution. Considering that the ocean surface is highly non-stationary, observing it from different angles might bring additional information about the illuminated area. Thus, we decompose the vignette into subapertures, each one corresponding to the image formed using only a part of the total azimuth angle. Decomposing the vignette, we can mimic different observation angles of the same scene, gathering more information. The SD algorithm is visually described in in the first part of the Fig. \ref{fig_pipeline}.

\noindent
\textbf{$\boldsymbol{\sigma_0}$ calibration.}
According to \cite{Wang-2019-GDJ}, the measured normalized radar cross section $\sigma_0$ by SAR over the ocean is highly dependent on the local ocean surface wind and viewing angles (incidence and azimuth) of the radar. Therefore, the $\sigma_0$ of each input vignette is calibrated by dividing it to a reference factor, constructed by assuming a constant wind of $10$ m/s at $45^{\circ}$ relative to the antenna look angle.

\noindent
\textbf{Azimuth FFT.}
The output of the $\sigma_0$ calibration block is fed into the azimuth Fast Fourier Transform (FFT) block, where we perform the FFT along the azimuth axis to obtain the vignette's spectrum. The number of FFT points is equal to the number of points in the azimuth direction.

\noindent
\textbf{Hamming window compensation.}
The spectrum composed by the azimuth FFT block is compensated with a Hamming window, having a coefficient of $0.75$, in order to obtain a flat azimuth spectrum. The result is shown in the second and third picture from Fig. \ref{fig_pipeline} left.

\noindent
\textbf{Subaperture generation.}
In the following stage, we filter the processed vignette with $4$ shifted Hamming windows (with the same $0.75$ coefficient), in order to obtain the corresponding azimuth spectrum for each subaperture.

\noindent
\textbf{Azimuth iFFT.}
Having the azimuth spectrum for each subaperture, we want to transform back the data into time domain by computing an inverse Fast Fourier Transform (iFFT), with the same parameters from the azimuth FFT block.
The time domain subapertures are forward processed by the DCE pipeline.

\begin{figure*}[!t]
\begin{center}
\centerline{\includegraphics[width=.87\linewidth]{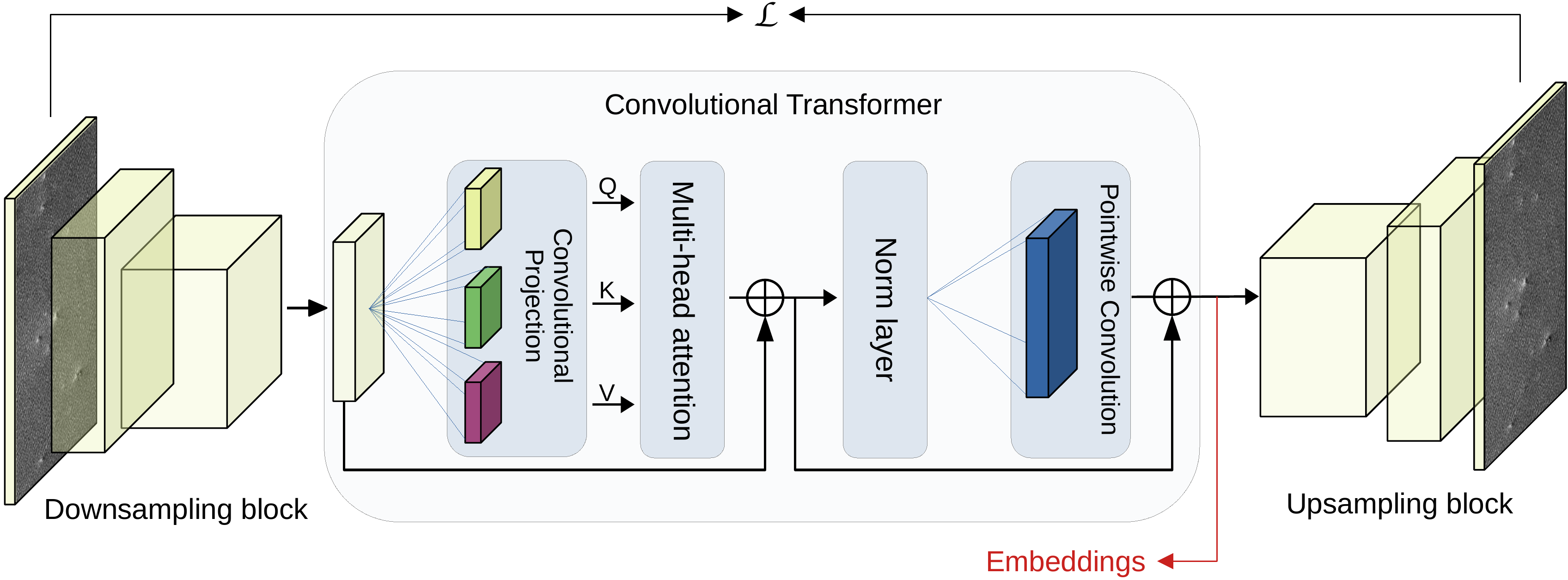}}
% \vspace{-0.25cm}
\caption{CyTran generative architecture. The model is formed of a downsampling block comprising convolutional layers, a convolution transformer block comprising a multi-head self-attention mechanism, and an upsampling block comprising transposed convolutions. By $\mathcal{L}$ we denote the mean square error loss function between the input and the output and with red arrow is illustrated the place where the descriptive embeddings are taken.}
\label{fig_unsuper}
%\vspace{-0.9cm}
\end{center}
\end{figure*}

\subsection{Doppler centroid estimation}

Let $X_i \in \mathbb{R}^{m \times n}$ be the $i^{th}$ subaperture for a vignette, where $m, n \in \mathbb{N}$ are the azimuth and range dimensions. Let $Y_i \in \mathbb{R}^{m \times n}$ be the delayed version with 1 sample in the azimuth axis of $X_i$. We estimate the Doppler centroids for the $i^{th}$ subaperture as follows:

\begin{equation}
    D_i = -PRF \cdot \frac{angle(Z_i)}{2 \pi},
\end{equation}
where $Z_i = filt(X_i \cdot Y_i^*)$, $Y_i^*$ is the complex conjugate of $Y_i$, $PRF$ is the pulse repetition frequency, $angle()$ returns the angle of the complex input and $filt()$ is a two dimensional mean filter with $d_1 \times d_2$ kernel size. Each estimated $D_i$ is further decimated. An illustration could be observed in the first two blocks of the Fig. \ref{fig_pipeline}.

\noindent
\textbf{Decimation.}
The last stage of the preprocessing pipeline is the decimation. The fine‐resolution subapertures or DCE are not necessary for large scale geophysical phenomena, especially since the classes described in \cite{Wang-2019-GDJ} have scales of tens to thousands of metres. Therefore to better highlight larger feature patterns, we low-pass-filter each resulted $X_i$ and $D_i$ with a window of $10\times10$, each filter's coefficient being $0.01$. The resulted images are then decimated by $1/10$ yielding a pixel spacing of 50 meters. We highlight that the decimation is performed for both SD and DCE in accordance with the desired output.

\subsection{Deep neural networks for classification}
The success of the CNNs in image processing tasks \cite{Dhillon-2020-PAI} encouraged their introduction in remote sensing applications and SAR imagery \cite{Wang-2019-RSE, Quach-2020-TRGS, Colin-2021-IGARSS, De-2021-JSTAR}. Thus, we followed our previous work \cite{Ristea-2022-IGARSS}, proposing a data-centring approach, rather than a novel model architecture. We focused our attention on the preprocessing stage and employed two well-known architectures, ResNet18 \cite{He-CVPR-2016} and InceptionV3 \cite{Szegedy-CVPR-2016}, for the ocean SAR image classification task. The networks were pretrained on the ImageNet data set and minimal architectural changes were made: the number of output neurons and the number of input channels.

\subsection{Unsupervised neural network}

In our work we used the CyTran generative architecture formed of a convolutional downsampling block, a convolutional transformer block, and a deconvolutional upsampling block, as illustrated in Fig. \ref{fig_unsuper}. We underline that, without the convolutional downsampling block and the replacement of dense layers with convolutional layers inside the transformer block, the transformer would not be able to learn to generate images larger than $64\times 64$ pixels, due to memory overflow (measured on a Nvidia GeForce RTX 3090 GPU with 24GB of VRAM).

The downsampling block starts with a convolutional layer formed of $32$ filters with a spatial support of $7 \times 7$, which are applied using a padding of $3$ pixels to preserve the spatial dimension, while enriching the number of feature maps to $32$. Next, we apply three convolutional layers composed of $32$, $64$ and $128$ filters, respectively. All convolutional filters have a spatial support of $3 \times 3$ and are applied at a stride of $2$, using a padding of $1$. Each layer is followed by batch-norm \cite{Ioffe-ICML-2015} and Rectified Linear Units (ReLU) \cite{Nair-ICML-2010}. 
The downsampling block is followed by the convolutional transformer block, which provides an output tensor of the same size as the input tensor. The convolutional transformer block is inspired by the block proposed in \cite{Ristea-2021-ARXIV}. More precisely, the input tensor is interpreted as a set of overlapping visual tokens. The sequence of tokens is projected onto a set of weight matrices implemented as depthwise separable convolution operations. The convolutional projection is formed of three nearly identical projection blocks, with separate parameters. The output query, keys and values are passed to a multi-head attention layer, with the goal of capturing the interaction among all tokens by encoding each entity in terms of the global contextual information. Next, the output passes through a batch-norm and a pointwise convolution. Lastly, the result of the convolutional transformer block is processed by the upsampling block, being designed to revert the transformation of the downsampling block.

We use CyTran architecture in an unsupervised manner, aiming for the identity function by performing input auto-encoding. Specifically, we want to exactly reproduce the input data, by optimizing the following loss function between the input $X$ and output $\hat{X}$.
\begin{equation}
    \mathcal{L} = (X - \hat{X})^2
\end{equation}

Finally, after the unsupervised training procedure, we use the CyTran model to encode into embeddings the input data for image retrieval. The embeddings are taken after the convolutional transformer block, as depicted in Fig. \ref{fig_unsuper} with a red arrow.

% \vspace{+0.6cm}

\begin{algorithm}[!t]
\caption{Physics guided content based SAR image retrieval}\label{alg_1}
\textbf{Input: }{$DB$ - database with SAR images; $(X)$ - samples from $DB$; $Q$ - query SAR image; $N_{max}$ -  the number of images returned; $\eta$ - learning rate; $\mathcal{L}$ - loss function; $d$ - cosine similarity.}\\
\textbf{Notations: }{$f$ - the CyTran model; $\hat{f}$ - embedding function from the CyTran model; $\theta$ - the weights of the model; \emph{sort} - a function that jointly sorts the input set; $\mathcal{N}(0, \Sigma)$ - the normal distribution of mean $0$ and standard deviation $\Sigma$}; $\mathcal{U}$ - uniform distribution\\
\textbf{Initialization: }{$\theta^{(0)} \sim \mathcal{N}(0, \Sigma)$}\\
\textbf{Output: }{$U$ - a set of $N_{max}$ elements from $DB$ similar to the query SAR image $Q$.}
\vspace{0.3em}
\algrule
\vspace{0.3em}
\textbf{Stage 1:}{ Unsupervised pre-training of auto-encoding model}
\vspace{0.3em}
\algrule
\begin{algorithmic}[1]
\For{$i \gets 1$ to $n$}
    \State $t \gets 0$
    \While{converge criterion not met}
     \State $X^{(t)} \gets$ mini-batch $\sim \mathcal{U}(DB)$
     \State $\theta^{(t+1)}_{i} = \theta^{(t)}_{i} - \eta^{(t)} \nabla{\mathcal{L}\left(\theta^{(t)}_{i}, X^{(t)} \right)}$
     \State $t \gets t + 1$
    \EndWhile
\EndFor
\vspace{0.2em}
\algrule
\vspace{0.2em}
\noindent\hspace{-2em}
\textbf{Stage 2:}{ Processing the database}
\vspace{0.2em}
\algrule
\vspace{0.2em}
\State $\hat{DB}$ = \textit{empty}
\For{$X \gets DB$}
    \State $\hat{X} = \hat{f}(X)$
    \State $\hat{DB} \gets (X, \hat{X})$
\EndFor
\vspace{0.2em}
\algrule
\vspace{0.2em}
\noindent\hspace{-2em}
\textbf{Stage 3:}{ SAR image retrieval}
\vspace{0.2em}
\algrule
\vspace{0.2em}
\State $D$ = \textit{empty}
\State $\hat{Q} = \hat{f}(Q)$
\For{$(X, \hat{X}) \gets \hat{DB}$}
    \State $m \gets d(\hat{Q}, \hat{X})$
    \State $D \gets (X, m)$
\EndFor   
\State $D \gets sort(D)$ with respect to $m$
\State $U \gets D[1:N_{max}]$

\end{algorithmic}
\end{algorithm}

\subsection{Content based image retrieval}

Considering a very large database with ocean SAR images, we propose an unsupervised algorithm which can find similar vignettes, serving researchers as a tool to study physical phenomena on the ocean surface. We formally describe the steps in the Algorithm~\ref{alg_1}.

We consider as requested input the database, the query image and some hyper-parameters. In the first stage, we train in an unsupervised fashion the CyTran auto-encoder model denoted by $f$. We optimize the model such that we obtain a close reconstruction of the input $X$. In the next stage, we remove the upsampling block from the CyTran model and we define by $\hat{f}$ the pretrained model that computes the descriptive embeddings for each vignette. Next, we process each SAR image from the database, associating in $\hat{DB}$ a pair formed by the original image $X$ and the corresponding embedding $\hat{X}$ vector. At the end of the stage two, we will have an associated embedding for each vignette. We highlight that, Stage 1 and Stage 2 must be performed only once and do not introduce any time overhead in the retrieval stage.
Lastly, in the Stage 3 we perform the actual image retrieval. We compute the embedding vector $\hat{Q}$ of the query image $Q$ by $\hat{Q} = \hat{f}(Q)$. Afterwards, we calculate the cosine distance between $\hat{Q}$ and each embedding vector from $\hat{DB}$, as follows:

\begin{equation}
    d(\hat{Q}, \hat{X}) = \frac{\hat{Q} \cdot \hat{X}}{ ||\hat{Q}|| \cdot ||\hat{X}||},
\end{equation}
where $||x||$ stands from the $L_2$-norm of vector $x$.

All the distances $m$ associated with the vignettes from $DB$ are cashed in $D$. In the last two steps from Stage 3, we sort $D$ in accordance with the distance $m$ and take the most $N_{max}$ similar examples. In this manner, we can obtain an arbitrary $N_{max}$ number of the most similar examples, by only computing the distance between the query embedding vector and sorting the result.

We emphasis that our algorithm is general, does not require labels and is not constrained for any specific input data. To demonstrate the generality of our method, we considered as input two distinct distribution data, the SAR subapertures and the Doppler centroids estimated from subapertures. Therefore, we perform a content based image retrieval from both raw data and physics aware representations. The algorithm feed with the latter data type could build a more complex search engine, capable to find phenomena based on specific physical features (e.g., ocean currents with a certain speed).

\begin{table*}[t]
\setlength{\tabcolsep}{1.6pt}
\renewcommand{\arraystretch}{1.6}
\centering
\noindent
\caption{Retrieval results on TenGeoPSAR-wv test set considering the embeddings from ResNet18 (S - supervised training) and CyTran (U - unsupervised training) models. We reported results when we consider as input data the original vignette (Vig) and all subapertures (Subap). By P@$m$ we denote the precision score for the most similar $m$ samples.} 
\begin{tabular}{|l|cc|cc|cc|cc|cc|cc|cc|cc|cc|cc|cc|}
\hline  
\multirow{2}{*}{Method} & 
\multicolumn{2}{c|}{{POW}} &
\multicolumn{2}{c|}{{WS}} &
\multicolumn{2}{c|}{{MCC}} &
\multicolumn{2}{c|}{{RC}} &
\multicolumn{2}{c|}{{BS}} &
\multicolumn{2}{c|}{{SI}} &
\multicolumn{2}{c|}{{Ic}} &
\multicolumn{2}{c|}{{LWA}} &
\multicolumn{2}{c|}{{AF}} &
\multicolumn{2}{c|}{{OF}}  & 
\multicolumn{2}{c|}{{Overall}} \\

& \rotatebox{90}{P@5} & \rotatebox{90}{P@50} & \rotatebox{90}{P@5} & \rotatebox{90}{P@50} & \rotatebox{90}{P@5} & \rotatebox{90}{P@50} & \rotatebox{90}{P@5} & \rotatebox{90}{P@50} & \rotatebox{90}{P@5} & \rotatebox{90}{P@50} & \rotatebox{90}{P@5} & \rotatebox{90}{P@50} & \rotatebox{90}{P@5} & \rotatebox{90}{P@50} & \rotatebox{90}{P@5} & \rotatebox{90}{P@50} & \rotatebox{90}{P@5} & \rotatebox{90}{P@50} & \rotatebox{90}{P@5} & \rotatebox{90}{P@50}  & \rotatebox{90}{P@5} & \rotatebox{90}{P@50} \\

\hline
S-Vig    & 100 &  99.8 &  99.8 &  99.7 &  100 &  99.6 &  99.4 &  98.5 &  99.4 &  98.9 &  99.8 &  99.7 &  97.2 &  96.4 &  97.2 &  96.9 &  97.4 &  94.8 &  91.6 &  89.4 & 98.1 & 97.4\\
\hline
% S-Vig    &  98.9 & 97.9 & 98.5 & 98.4 & 100 & 99.7 & 97.6 & 96.3 & 95.7 & 95.6 & 98.5 & 97.4 & 95.6 & 94.5 & 97.0 & 96.7 & 95.2 & 93.3 & 89.5 & 88.4 & 99.1 & 99.1 \\ 
\hline
S-Subap    & 99.6 &  99.8 &  99.2 &  99.5 &  99.2 &  98.8 &  98.2 &  96.5 &  99.8 &  99.7 &  99.4 &  98.9 &  99.4 &  98.1 &  98.2 &  96.7 &  98.8 &  94.4 &  97.2 &  89.6 & 98.9 & 97.2\\ 
\hline
\hline

U-Vig    & 76.8 &  64.0 &  46.0 &  32.1 &  37.6 &  22.1 &  46.6 &  28.8 &  49.0 &  33.3 &  38.4 &  22.7 &  30.0 &  11.8 &  89.8 &  84.5 &  33.8 &  17.2 &  26.6 &  9.3  & 47.4 & 32.6\\ 

\hline
U-Subap    & 89.8 &  83.2 &  82.2 &  70.9 &  64.2 &  51.6 &  57.0 &  38.9 &  78.6 &  66.9 &  76.0 &  54.2 &  57.0 &  38.0 &  91.0 &  82.1 &  63.8 &  47.2 &  66.8 &  39.6 & 72.6 & 57.3\\ 

\hline

\end{tabular}
\label{tab_subap_u}
%\vspace{-0.3cm}
\end{table*}

\section{Experimental setup}

\noindent
\textbf{Data set.}
TenGeoP‐SARwv data set contains over 37,000 ocean vignettes with 10 geophysical phenomena. Following \cite{Ristea-2022-IGARSS}, we used the raw vignettes from the TenGeoP‐SARwv data set, with the assigned labels, and randomly split the data in training (70\%), validation (15\%) and test (15\%). Moreover, for Doppler based experiments we processed the raw vignettes in accordance with the full data pipeline described in Fig. \ref{fig_pipeline}. 
Further, for brevity we will use the following abbreviations for data set classes: POW - Pure Ocean Waves, WS - Wind Streaks, MCC - Micro Convective Cells, RC - Rain Cells, BS - Biological Slicks, SI - Sea Ice, Ic - Iceberg, LWA - Low Wind Area, AF - Atmospheric Front, OF - Oceanic Front.

\noindent
\textbf{Hyper-parameters tuning.}
For the classification experiment, we tuned the hyper-parameters similar to \cite{Ristea-2022-IGARSS}. Regarding the CyTran model, we used the same network hyper-parameters as proposed in \cite{Ristea-2021-ARXIV}, only adjusting the input and output number of channels, in accordance with the input type. We trained the model for $100$ epochs using Adam optimizer and a mini-batch size of 16. Regarding DCE, we used $d_1=d_2=32$ for the mean filter.

\noindent
\textbf{Evaluation metrics.}
We reported the accuracy for the classification task and performed McNemar statistic tests to show the statistical significance of our results. Regarding the retrieval task, considering that we target big data streams, we reported the precision for $5$ ($P@5$) and $50$ ($P@50$) examples. Each score was averaged for $100$ queries, more precisely, we computed $P@5$ and $P@50$ for 100 query samples and averaged the results.

\begin{figure}[!t]
\begin{center}
\centerline{\includegraphics[width=1\linewidth]{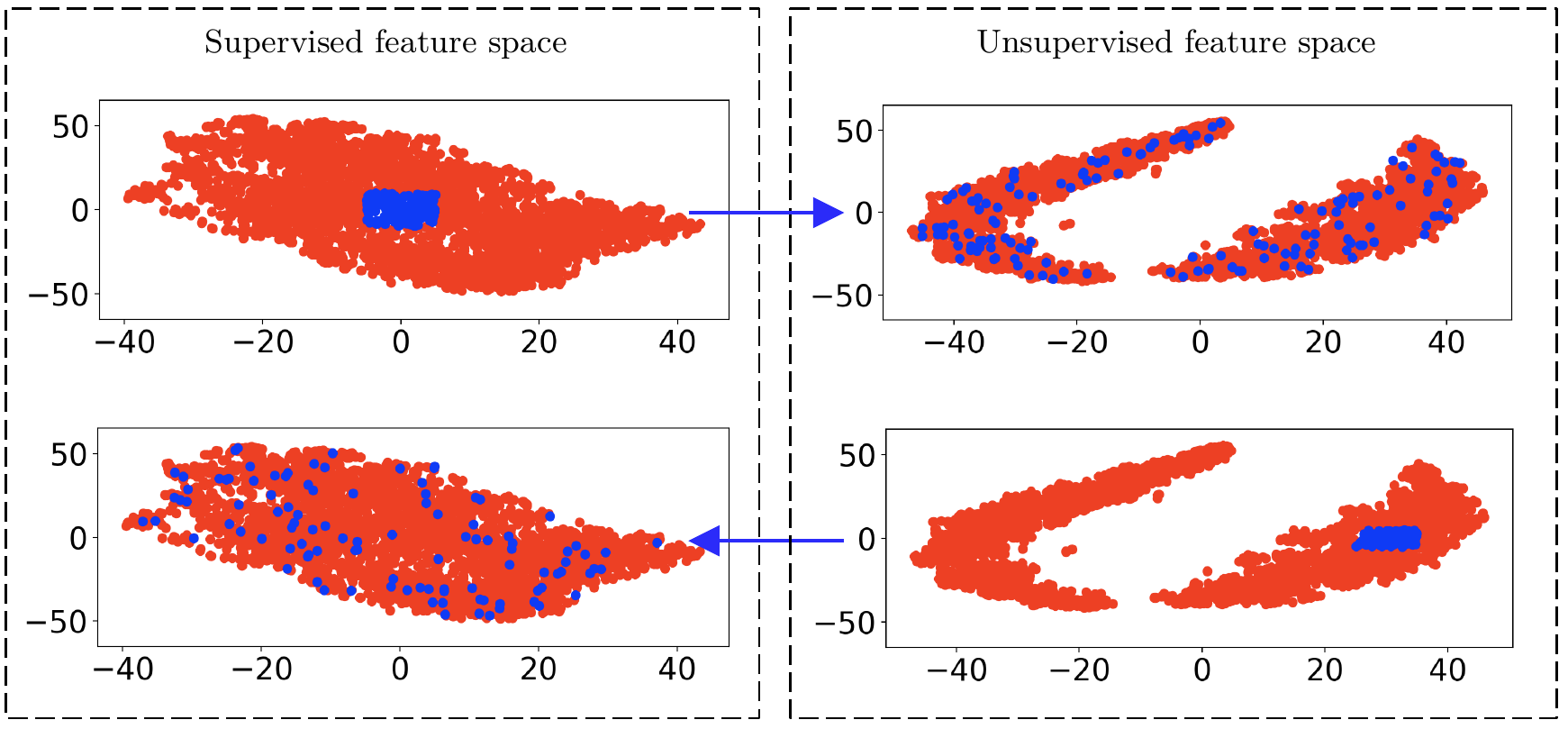}}
% \vskip -0.3cm
\caption{Embedding space comparison for the biological slicks class between supervised and unsupervised trainings. The figures are horizontally corespondent, indicating the samples annalogy between feature spaces. The dimensional reduction is computed with T-SNE.}
\label{fig_BS}
\end{center}
% \vskip -1.0cm
\end{figure}

\begin{figure}[!t]
\begin{center}
\centerline{\includegraphics[width=1\linewidth]{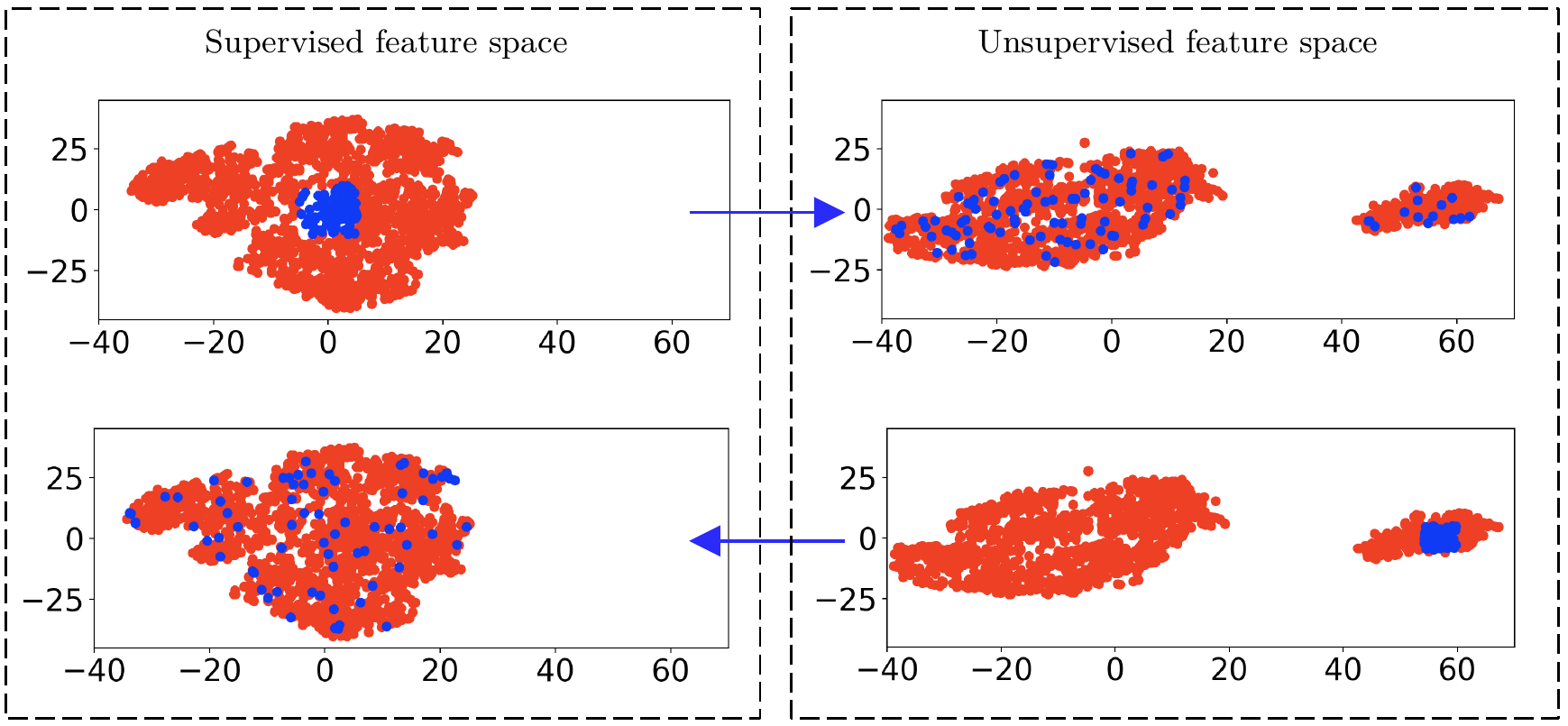}}
% \vskip -0.3cm
\caption{Embedding space comparison for the low wind area class between supervised and unsupervised trainings. The figures are horizontally corespondent, indicating the samples annalogy between feature spaces. The dimensional reduction is computed with T-SNE.}
\label{fig_LWA}
\end{center}
% \vskip -1.0cm
\end{figure}

\begin{table}
\centering
\caption{Accuracy results for a ResNet18 model on the TenGeoP‐SARwv test set. We denote by ``Subaperture (1)'' that the input is only the first subaperture, while for ``Subapertures'' all four are considered. The significantly better results (level $0.01$) than corresponding baselines, according to a paired McNemar's test, are marked with $\dagger$.}

\label{tab_results_classification}
 \begin{tabular}{|l|c|}

 \hline
 Vignette                   & 98.0    \\ 
 \hline
 Subaperture (1)               & 94.0      \\
 \hline
 Subapertures         & 98.9$^{\dagger}$     \\
 \hline
 \hline
 DCE Vignette                     & 78.6     \\ 
 \hline
 DCE Subapertures (1)                & 75.3       \\
 \hline
 DCE Subapertures          & 93.3$^{\dagger}$    \\
 \hline

\end{tabular}
\end{table}

\begin{figure*}[!t]
\begin{center}
\centerline{\includegraphics[width=1\linewidth]{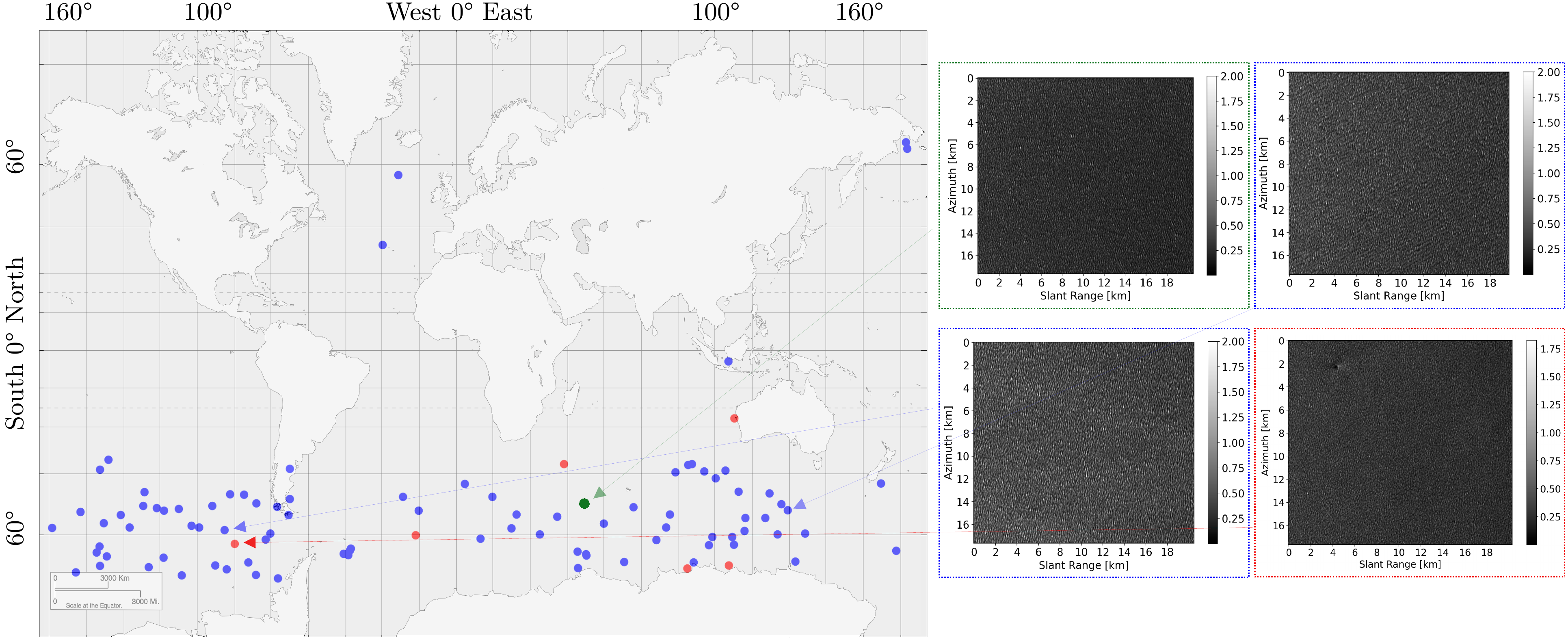}}
% \vspace{-0.25cm}
\caption{Retrieval results based on embeddings from CyTran model trained on all subapertures from the original vignette. We present the most similar $N_{max}=100$ samples with localisation information. In green is represented the query image, in blue the images found from the same class and in red the images found from wrong classes. In the right side, we show the original vignette for some samples: green and blue (Pure Ocean Waves), red (Iceberg).}
\label{fig_earth_POW}
\vspace{-0.3cm}
\end{center}
\end{figure*}

\begin{figure*}[!t]
\begin{center}
\centerline{\includegraphics[width=1\linewidth]{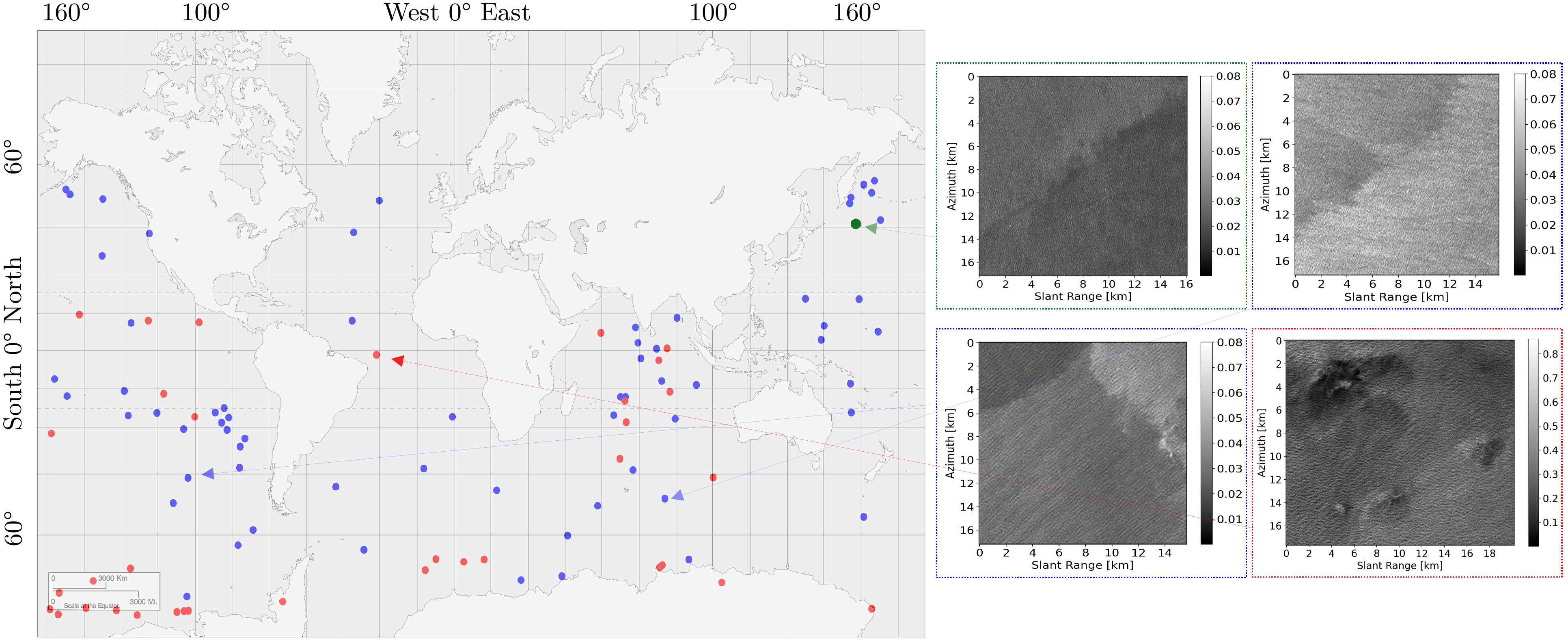}}
% \vspace{-0.25cm}
\caption{Retrieval results based on embeddings from CyTran model trained on all subapertures from the original vignette. We present the most similar $N_{max}=100$ samples with localisation information. In green is represented the query image, in blue the images found from the same class and in red the images found from wrong classes. In the right side, we show the original vignette for some samples: green and blue (Atmospheric Front), red (Micro Convective Cells).}
\label{fig_earth_AF}
\vspace{-0.2cm}
\end{center}
\end{figure*}

\begin{table*}[!t]
\setlength{\tabcolsep}{1.6pt}
\renewcommand{\arraystretch}{1.6}
\centering
\noindent
\caption{Retrieval results on TenGeoPSAR-wv test set considering the embeddings from ResNet18 (S - supervised training) and CyTran (U - unsupervised training) models. We reported results when we consider as input data the DCE on the original vignette (Vig) and DCE on all subapertures (Subap). By P@$m$ we denote the precision score for the most similar $m$ samples.} 
\begin{tabular}{|l|cc|cc|cc|cc|cc|cc|cc|cc|cc|cc|cc|}
\hline  
\multirow{2}{*}{Method} & 
\multicolumn{2}{c|}{{POW}} &
\multicolumn{2}{c|}{{WS}} &
\multicolumn{2}{c|}{{MCC}} &
\multicolumn{2}{c|}{{RC}} &
\multicolumn{2}{c|}{{BS}} &
\multicolumn{2}{c|}{{SI}} &
\multicolumn{2}{c|}{{Ic}} &
\multicolumn{2}{c|}{{LWA}} &
\multicolumn{2}{c|}{{AF}} &
\multicolumn{2}{c|}{{OF}}  & 
\multicolumn{2}{c|}{{Overall}} \\

& \rotatebox{90}{P@5} & \rotatebox{90}{P@50} & \rotatebox{90}{P@5} & \rotatebox{90}{P@50} & \rotatebox{90}{P@5} & \rotatebox{90}{P@50} & \rotatebox{90}{P@5} & \rotatebox{90}{P@50} & \rotatebox{90}{P@5} & \rotatebox{90}{P@50} & \rotatebox{90}{P@5} & \rotatebox{90}{P@50} & \rotatebox{90}{P@5} & \rotatebox{90}{P@50} & \rotatebox{90}{P@5} & \rotatebox{90}{P@50} & \rotatebox{90}{P@5} & \rotatebox{90}{P@50} & \rotatebox{90}{P@5} & \rotatebox{90}{P@50}  & \rotatebox{90}{P@5} & \rotatebox{90}{P@50} \\

\hline
S-Dop vig    & 88.6 &  86.3 &  70.2 &  62.2 &  67.2 &  55.1 &  87.6 &  83.6 &  89.2 &  87.5 &  94.6 &  92.6 &  67.6 &  54.4 &  95.2 &  94.9 &  65.2 &  54.2 &  43.6 &  27.2 & 76.9  & 69.8 \\ 
\hline
S-Dop Subap    & 96.2 &  94.64 &  94.2 &  92.0 &  97.0 &  96.4 &  96.2 &  95.5 &  98.4 &  97.0 &  97.6 &  97.0 &  81.6 &  73.3 &  98.2 &  97.1 &  91.8 &  89.8 &  62.6 &  47.8 & 91.3 & 88.0\\ 

\hline
\hline

U-Dop vig    & 84.2 &  76.3 &  58.0 &  42.8 &  43.2 &  27.3 &  43.4 &  23.6 &  61.2 &  46.1 &  47.0 &  26.7 &  32.6 &  17.2 &  79.0 &  73.5 &  35.8 &  16.9 &  27.4 &  9.6 & 51.1 & 36.0\\ 

\hline
U-Dop Subap    & 90.4 &  83.5 &  74.4 &  63.1 &  59.4 &  47.0 &  55.4 &  40.3 &  75.0 &  64.8 &  61.4 &  42.9 &  55.6 &  36.2 &  85.2 &  69.2 &  58.8 &  36.8 &  52.2 &  36.0 & 66.7 & 52.0 \\ 

\hline

\end{tabular}
\label{tab_dopp_u}
%\vspace{-0.3cm}
\end{table*}

\section{Experimental results}

\noindent
\textbf{Classification results.}
We extend the results from \cite{Ristea-2022-IGARSS} in Table \ref{tab_results_classification}, where we report the classification accuracy obtained for the ResNet18 model on TenGeoP‐SARwv test set, considering multiple inputs data types. When we consider as training input all the subapertures computed on the vignette, we observe a performance boost of $0.9\%$, with respect to the model trained on the original vignette. But, when we feed only the first subaperture, an accuracy drop of $4\%$ occurs. Similarly, when we train the model on DCE on subapertures against DEC on original vignette, we observe a drastically improvement of $14.7\%$. This highlights that the SD algorithm applied on the ocean vignettes helps the training process for both raw SAR data and DCE. 

\noindent
\textbf{Unsupervised training results.}
We trained the CyTran \cite{Ristea-2021-ARXIV} auto-encoder model on the TenGeoP‐SARwv training set and choose the best model with respect to the reconstruction loss on the evaluation set. We highlight that multiple models were tried (e.g., ResNet auto-encoder, U-Net) but they did not converge to optimal reconstruction results, therefore we excluded them.

We visualized with T-SNE the embedding feature space when we considered as input all the subapertures on the original vignette for both the supervised trained model and CyTran. For a more accurate comparison, we did the visualization class by class and included the results for biological slicks, in Fig. \ref{fig_BS}, and low wind area, in Fig. \ref{fig_LWA}. We note that the feature space is distinct (for CyTran embeddings we clearly see two cloud points for both figures), suggesting that into the same annotated class from TenGeoP‐SARwv we could find distinguishable phenomena. Moreover, the distance metric is not preserved between feature spaces, emphasized by the blue points from both Fig. \ref{fig_BS} and Fig. \ref{fig_LWA}, which are close in one feature space and randomly spread into the other.

\noindent
\textbf{Retrieval results.}
On the one hand, in Table \ref{tab_subap_u} we reported the retrieval performance on embeddings provided by CyTran network, trained on original vignette and subapertures. We compared the retrieval results against the embeddings computed by ResNet18 model trained in a supervised fashion. When we compare the supervised embeddings on original vignette (S-Vig) and subapertures (S-Subap), the results are comparable, with overall differences smaller than $1\%$ for both $P@5$ and $P@50$. But, the SD algorithm offers a consistent precision boost when we refer to the retrieval results with unsupervised embeddings. We observe that the unsupervised embeddings on subapertures raise the $P@5$ and $P@50$ for each and every class, with an overall improvement of $25.2\%$ for $P@5$ and $24.7\%$ for $P@50$. Even if, the SD does not bring a precision boost for supervised embeddings, most probably because of the saturated accuracy on the data set, the algorithm has a huge impact in unsupervised scenarios, reducing the retrieval performance gap between supervised and unsupervised approaches.

On the other hand, we did a more physics based experiment, considering as input data the DCE, which are directly correlated with physic phenomena (e.g., ocean currents). In Table \ref{tab_dopp_u} we reported the retrieval performance on embeddings provided by CyTran network, trained on DCE on original vignette and subapertures. We compared the retrieval results against the embeddings computed by ResNet18 model trained in a supervised fashion. As we would expect from the classification experiment, the retrieval performance is considerable improved when the supervised embeddings based on subapertures are used. The same trend is observed for the unsupervised embeddings. More precisely, the $P@5$ for U-Dop Subap is with $15.6\%$ higher than U-Dop Vig and the $P@50$ is with $16.0\%$ higher. Thus, SD algorithm has a major positive impact on the retrieval task, when DCE data are used, leading the way to more complex search engines.

Additionally, we showed the retrieval results for the unsupervised embeddings trained on subapertures for two query images: in Fig. \ref{fig_earth_POW} for pure ocean waves class and in Fig. \ref{fig_earth_AF} for atmospheric front class. For both figures, we observe that the most similar images found are randomly spread in the geographical area where the phenomena could appear, indicating that the unsupervised model does not overfit with respect to the geographical area. Moreover, structural similarities were observed for the images found with wrong label (the red points from Fig. \ref{fig_earth_POW} and Fig. \ref{fig_earth_AF}), which can indicate the presence of two phenomena in the same location or other intrinsic similarities.

\section{Conclusion}
In this work, we extended the previous approach from \cite{Ristea-2022-IGARSS} by using the SD algorithm for unsupervised feature learning with transformer networks. The unsupervised features were used for a SAR retrieval algorithm on the ocean surface, showing important improvements in performance when the SD was used as a pretraining stage for the models. Moreover, we showed that the SD method has a huge impact in retrieval performance when more physics based algorithms, as DCE, are used for ocean retrieval. This experiment allow us to build more complex searching engines, which could find similar physical parameters, instead of similar structures (e.g., ocean currents speed). Summing up, we used a data-centring approach to improve the performance classification and retrieval algorithms, in both supervised and unsupervised settings.

\section*{Acknowledgment}
This work was supported by IFREMER and by the grant of the Romanian Ministry of Education and Research, CNCS - UEFISCDI, project number PN-III-P4-ID-PCE-2020-2120, within PNCDI III.

% \ifCLASSOPTIONcaptionsoff
%   \newpage
% \fi

\bibliographystyle{IEEEtran}
\bibliography{report}

\end{document}